%% file: arxiv.tex
\documentclass[runningheads]{llncs}
\usepackage{graphicx}
\usepackage{comment}
\usepackage{amsmath,amssymb} 
\usepackage{color}

\usepackage[width=122mm,left=12mm,paperwidth=146mm,height=193mm,top=12mm,paperheight=217mm]{geometry}

\usepackage{epsfig}
\usepackage{xspace}
\usepackage{multirow}
\usepackage{colortbl}
\usepackage{tabularx}
\usepackage{booktabs}
\usepackage{caption}
\usepackage{tablefootnote}
\usepackage{adjustbox}
\usepackage{enumitem}
\usepackage[normalem]{ulem}
\usepackage[table]{xcolor}
\usepackage{footmisc}

\usepackage[pagebackref=true,breaklinks=true,colorlinks,bookmarks=false]{hyperref}
\usepackage{xr}

\usepackage{enumitem}
\setitemize{noitemsep,topsep=0pt,parsep=0pt,partopsep=0pt,leftmargin=*}

\newcommand{\iABNsync}{iABN$^\text{sync}$\xspace}

\newcommand{\convx}[1]{\textrm{conv#1}}
\newcommand{\monodis}{MonoDIS}

\newcommand{\vct}[1]{\ensuremath{\boldsymbol{#1}}}

\newcommand{\con}[1]{\ensuremath{\mathsf{#1}}}

\newcommand{\ie}{\textit{i.e.}\xspace}
\newcommand{\eg}{\textit{e.g.}\xspace}
\newcommand{\etc}{\textit{etc.}\xspace}

\newcommand{\MoVi}{\textit{MoVi-3D}\xspace}

\newcommand{\footnoteref}[1]{\textsuperscript{\ref{#1}}}

\definecolor{mapillarygreen}{RGB}{5,203,99}
\definecolor{darkgreen}{cmyk}{1,0,0.83,0}

\renewcommand{\paragraph}[1]{
        \vspace{2pt}
	\noindent\textbf{#1}}

\begin{document}
\pagestyle{headings}
\mainmatter
\def\ECCVSubNumber{4185}  

\title{Towards Generalization Across Depth for Monocular 3D Object Detection} 

%
\author{Andrea Simonelli\inst{1,2,3} \and
Samuel Rota Bul\'o\inst{1} \and
Lorenzo Porzi\inst{1} \and
Elisa Ricci\inst{2,3} \and
Peter Kontschieder\inst{1}}
\authorrunning{A. Simonelli et al.}

\institute{Mapillary Research, Graz, Austria \\
\url{https://research.mapillary.com} \and
University of Trento, Trento, Italy \and
Fondazione Bruno Kessler, Trento, Italy}

\maketitle

\begin{abstract}
\vspace{-15pt}
While expensive LiDAR and stereo camera rigs have enabled the development of successful 3D object detection methods, monocular RGB-only approaches lag much behind. This work advances the state of the art by introducing \MoVi, a novel, single-stage deep architecture for monocular 3D object detection. \MoVi builds upon a novel approach which leverages geometrical information to generate, both at training and test time, virtual views where the object appearance is normalized with respect to distance. 
These virtually generated views facilitate the detection task as they significantly reduce the visual appearance variability associated to objects placed at different distances from the camera. As a consequence, the deep model is relieved from learning depth-specific representations and its complexity can be significantly reduced. In particular, in this work we show that, thanks to our virtual views generation process, a lightweight, single-stage architecture suffices to set new state-of-the-art results on the popular KITTI3D benchmark.
\vspace{-15pt}
\end{abstract}

\begin{figure}[t]
    \includegraphics[width=1.0\textwidth]{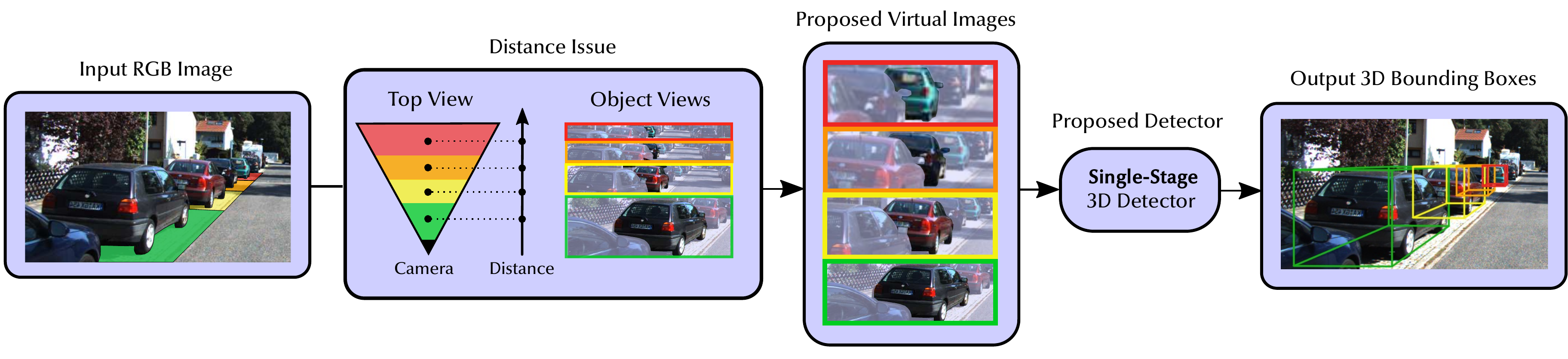}
    \caption{We aim at predicting a 3D bounding box for each object given a single image (left). In this image, the scale of an object heavily depends on its distance with respect to the camera. For this reason the complexity of the detection increases as the distance grows. Instead of performing the detection on the original image, we perform it on virtual images (middle). Each virtual image presents a cropped and and scaled version of the original image that preserves the scale of objects as if the image was taken at a different, given depth. Colors and masks have been used for illustrative purposes only.}
    \label{fig:teaser}
    \vspace{-20pt}
\end{figure}

\vspace{-10pt}
\section{Introduction}\label{sec:introduction}
\input{intro.tex}

\section{Related Work}
\input{related.tex}

\section{Problem Description}
\begin{figure}[b]
    \centering
    \vspace{-15pt}
    \includegraphics[width=.8\columnwidth]{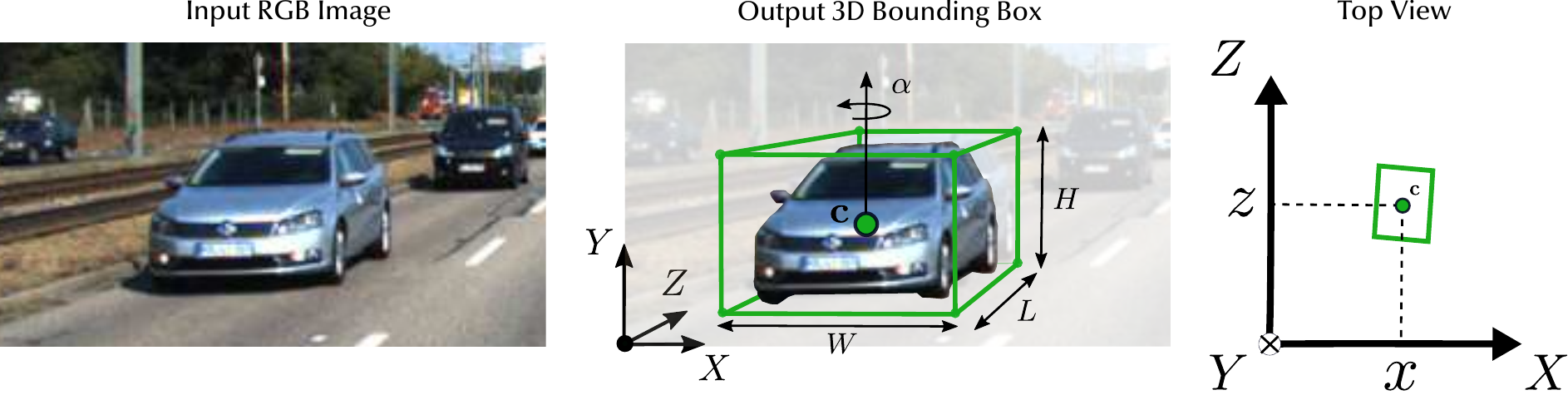}
    \caption{Illustration of the Monocular 3D Object Detection task. Given an input image (left), the model predicts a 3D box for each object (middle). Each box has its 3D dimensions $\textbf{s}=(W,H,L)$, 3D center $\textbf{c} = (x,y,z)$ and rotation ($\alpha$).}
    \label{fig:problem}
    \vspace{-10pt}
\end{figure}

In this work we address the problem of monocular 3D object detection, illustrated in Fig.~\ref{fig:problem}. Given a single RGB image, the task consists in predicting 3D bounding boxes and an associated class label for each visible object. The set of object categories is predefined and we denote by $\con n_c$ their total number.

 In contrast to other methods in the literature, our method makes no use of additional information such as pairs of stereo images, or depth derived from LiDAR or obtained from monocular depth predictors (supervised or self-supervised). In order to boost their performance, the latter approaches tend to use depth predictors that are pre-trained on the same dataset where monocular 3D object detection is going to be run.
Accordingly, the setting we consider is the hardest and in general ill-posed. The only training data we rely on consists of RGB images with annotated 3D bounding boxes. Nonetheless, we assume that per-image camera calibrations are available at both training and test time.

\vspace{-10pt}
\section{Proposed Method}\label{sec:vc}
A deep neural network that is trained to detect 3D objects in a scene from a single RGB image is forced to build multiple representations for the same object, in a given pose, depending on the distance of the object from the camera. This is, on one hand, inherently due to the scale difference that two objects positioned at different depths in the scene exhibit when projected on the image plane. On the other hand, it is the scale difference that enables the network to regress the object's depth. In other words, the network has to build distinct representations devoted to recognize objects at specific depths and there is a little margin of generalization across different depths.
As an example, if we train a 3D car detector by limiting examples in a range of maximum $20$m and then at test time try to detect objects at distances larger than $20$m, the detector will fail to deliver proper predictions. We conducted this and other similar experiments and report results in Tab.~\ref{tab:kitti_ablation_distance}, where we show the performance of a state-of-the-art method MonoDIS~\cite{simonelli2019disentangling} against the proposed approach, when training and validating on different depth ranges. Standard approaches tend to fail in this task as opposed to the proposed approach, because they lack the ability to generalize across depths. As a consequence, when we train the 3D object detector we need to scale up the network's capacity as a function of the depth ranges we want to be able to cover and scale up accordingly the amount of training data, in order to provide enough examples of objects positioned at several possible depths. 

Our goal is to devise a training and inference procedure that enables generalization across depth, by indirectly forcing the models to develop representations for objects that are less dependent on their actual depth in the scene. The idea is to feed the model with transformed images that have been put into a canonical form that depends on some query depth. To illustrate the idea, consider a car in the scene and assume to virtually put a 2D window in front of the car. The window is parallel to the image plane and has some pre-defined size in meters. Given an output resolution, we can crop a 2D region from the original image corresponding to the projection of the aforementioned window on the image plane and rescale the result to fit the desired resolution. After this transformation, no matter where the car is in space, we obtain an image of the car that is consistent in terms of the scale of the object. Clearly, depth still influences the appearance, \eg due to perspective deformations, but by removing the scale factor from the nuisance variables we are able to simplify the task that has to be solved by the model.
In order to apply the proposed transformation we need to know the location of the 3D objects in advance, so we have a chicken-egg problem. In the following, we will show that this issue can be easily circumvented by exploiting geometric priors about the position of objects while designing the training and inference stages.

\begin{figure}[t]
    \centering
    \includegraphics[width=.5\columnwidth]{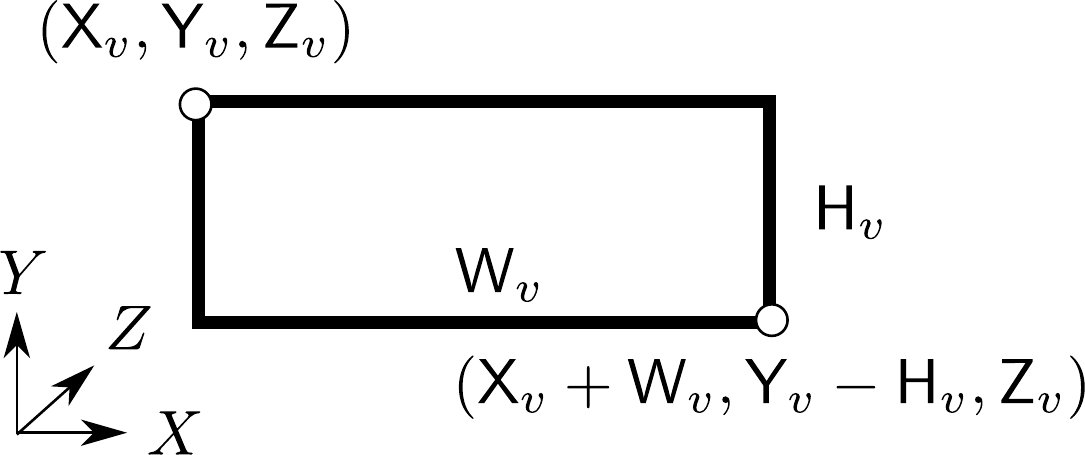}
    \caption{Notations about the 3D viewport.}
    \label{fig:vcam}
    \vspace{-15pt}
\end{figure}

\paragraph{Image transformation.} The proposed transformation is applied to the original image given a desired \emph{3D viewport}. A 3D viewport is a rectangle in 3D space, parallel to the camera image plane and positioned at some depth $\con Z_v$. The top-left corner of the viewport in 3D space is given by $(\con X_v, \con Y_v, \con Z_v)$ and the viewport has a pre-defined height $\con H_v$ thus spanning the range $[\con Y_v-\con H_v, \con Y_v]$ along the $Y$-axis (see Fig.~\ref{fig:vcam}).
We also specify a desired resolution $\con h_v\times\con w_v$ for the images that should be generated. 
%
The size $\con W_v$ of the viewport along the $X$-axis can then be computed as $\con W_v=\con w_v\frac{\con H_v}{\con h_v}\frac{f_y}{f_x}$, where $\con f_{x/y}$ are the $x/y$ focal lengths.
In practice, given an image captured with the camera and the viewport described above, we can generate a new image as follows.
We compute the top-left and bottom-right corners of the viewport, namely $(\con X_v, \con Y_v, \con Z_v)$ and $(\con X_v+\con W_v, \con Y_v-\con H_v, \con Z_v)$ respectively, and project them to the image plane of the camera, yielding the top-left and bottom-right corners of a \emph{2D} viewport. We crop it and rescale it to the desired resolution $\con w_v\times\con h_v$ to get the final output. We call the result a \emph{virtual image} generated by the given 3D viewport.

\begin{figure}[t]
    \centering
    \includegraphics[width=1.0\columnwidth]{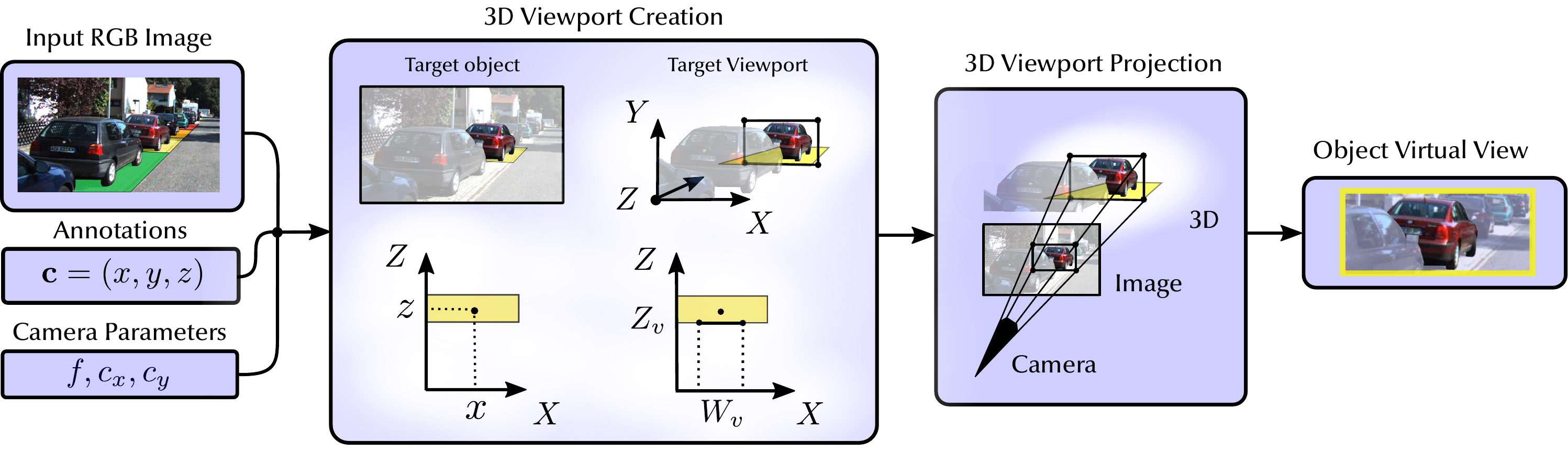}
    \vspace{-10pt}
    \caption{Training virtual image creation. We randomly sample a target object (dark-red car). Given the input image, object position and camera parameters, we compute a 3D viewport that we place at $z=Z_{v}$. We then project the 3D viewport onto the image plane, resulting in a 2D viewport. We finally crop the corresponding region and rescale it to obtain the target \textit{virtual view} (right). Colors and object masks have been used for illustrative purposes only.}
    \label{fig:train_scheme}
    \vspace{-20pt}
\end{figure}

\paragraph{Training.}
The goal of the training procedure is to build a network that is able to make correct predictions within a limited depth range given an image generated from a 3D viewport. Accordingly, we define a depth resolution parameter $\con Z_\text{res}$ that is used to delimit the range of action of the network. 
Given a training image from the camera and a set of ground-truth 3D bounding boxes, we generate $\con n_{v}$ virtual images from random 3D viewports. The sampling process however is not uniform, because objects occupy a limited portion of the image and drawing 3D viewports blindly in 3D space would make the training procedure very inefficient. Instead, we opt for a ground-truth-guided sampling procedure, where we repeatedly draw (without replacement) a ground-truth object and then sample a 3D viewport in a neighborhood thereof so that the object is completely visible in the virtual image. In Fig.~\ref{fig:train_scheme} we provide an example of such a sampling result.
The location of the 3D viewport is perturbed with respect to the position of the target ground-truth object in order to obtain a model that is robust to depth ranges up to the predefined depth resolution $\con Z_\text{res}$, which in turn plays an important role at inference time. Specifically, we position the 3D viewport in a way that $\con Y_v={\hat Y}$ and $\con Z_v={\hat Z}$, where $ {\hat Y}$ and $ {\hat Z}$ are the upper and lower bounds of the target ground-truth box along the $Y$- and $Z$-axis, respectively. From there, we shift $\con Z_v$ by a random value in the range $[-\frac{\con Z_\text{res}}{2},0]$, perturb randomly $\con X_v$ in a way that the object is still entirely visible in the virtual image and perturb $\con Y_v$ within some pre-defined range. 
The ground-truth boxes with $\hat Z$ falling outside the range of validity $[0,\con Z_\text{res}]$ are set to \emph{ignore}, \ie~there will be no training signal deriving from those boxes but at the same time we will not penalize potential predictions intersecting with this area. 
Our goal is to let the network focus exclusively on objects within the depth resolution range, because objects out of this range will be captured by moving the 3D viewport as we will discuss below when we illustrate the inference strategy.
Every other ground-truth box that is still valid will be shifted along the $Z$-axis by $-\con Z_v$, because we want the network to predict a depth value that is relative to the 3D viewport position. This is a key element to enforce generalization across depth.
In addition, we let a small share of the $\con n_v$ virtual images to be generated by 3D viewports randomly positioned in a way that the corresponding virtual image is completely contained in the original image. 
Finally, we have also experimented a class-uniform sampling strategy which allows to get an even number of virtual images for each class that is present in the original image.

\begin{figure}[th]
    \centering
    \includegraphics[width=\columnwidth]{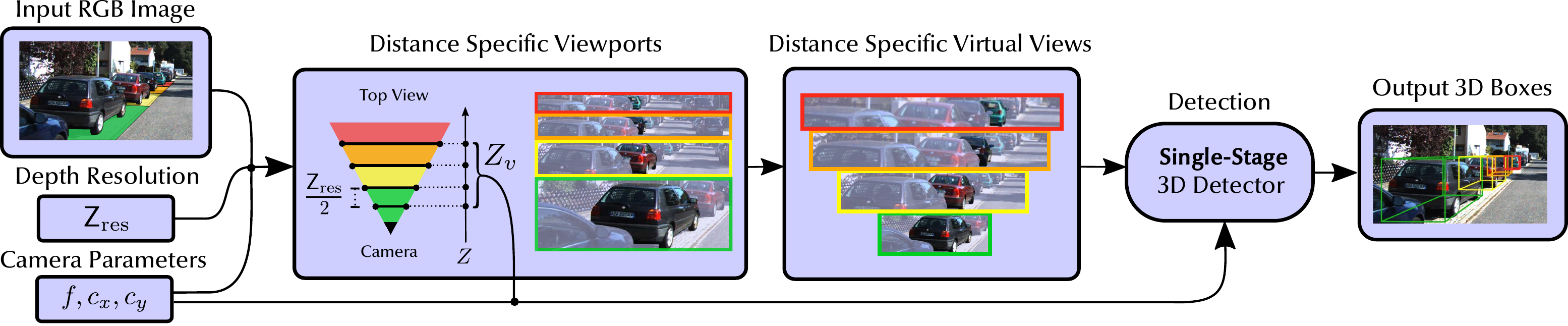}
    \caption{Inference pipeline. Given the input image, camera parameters and $\con Z_{\text{res}}$ we create a series of 3D viewports which we place every $\frac{\con Z_{\text{res}}}{2}$ meters along the $Z$ axis. We then project these viewports onto the image (as done during training, illustrated in Fig.~\ref{fig:train_scheme}), crop and rescale the resulting regions to obtain distance-specific \textit{virtual views}. We finally use these views to perform the 3D detection. Colors and object masks have been used for illustrative purposes only.}
    \label{fig:inference_vc}
    \vspace{-10pt}
\end{figure}

\paragraph{Inference.}
At inference time we would ideally put the 3D viewport in front of potential objects in order to have the best view for the detector. Clearly, we do not know in advance where the objects are, but we can exploit the special training procedure that we have used to build the model and perform a complete sweep over the input image by taking depth steps of $\frac{\con Z_\text{res}}{2}$ and considering objects lying close to the ground, \ie~we set $\con Y_v=0$. An illustration of the procedure in given in Fig~\ref{fig:inference_vc}. Since we have trained the network to be able to predict at distances that are twice the depth step, we are reasonably confident that we are not missing objects, in the sense that each object will be covered by at least a virtual image. Also, due to the convolutional nature of the architecture we adjust the width of the virtual image in a way to cover the entire extent of the input image. By doing so we have virtual images that become wider as we increase the depth, following the rule $\con w_v=\frac{\con h_v}{\con H_v}\frac{\con Z_v}{\con f_y}\con W$, where $\con W$ is the width of the input image. We finally perform NMS over detections that have been generated from the same virtual image.

\vspace{-10pt}
\section{Proposed Single-Stage Architecture}
We propose a \emph{single-stage}, fully-convolutional architecture for 3D object detection (\MoVi), consisting of a small backbone to extract features and a simple 3D detection head providing dense predictions of 3D bounding boxes. Details about its components are given below.

\vspace{-10pt}
\subsection{Backbone}
\begin{figure*}[thb]
    \centering
    \includegraphics[width=\textwidth]{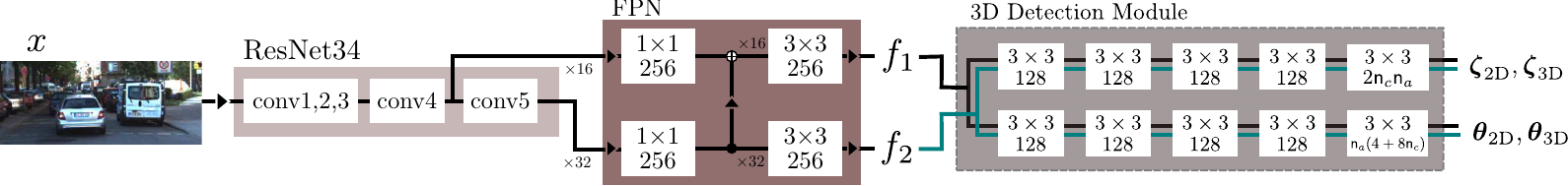}
    \caption{Our architecture. The backbone consists of a ResNet34 with a reduced FPN module covering only $2$ scales at $\times 16$ and $\times 32$ downsampling factors. The 3D detection head is run independently on $f_1$ and $f_2$. Rectangles in FPN and the 3D detection head denote convolutions followed by \iABNsync. See Sec.~\ref{ss:head} for a description of the different outputs.}
    \label{fig:architecture}
    \vspace{-10pt}
\end{figure*}

The backbone we adopt is a ResNet34~\cite{He2015b} with a Feature Pyramid Network (FPN)~\cite{Lin2016} module on top. The structure of the FPN network differs from the original paper~\cite{Lin+17} for we implement only 2 scales, connected to the output of modules \convx4 and \convx5 of ResNet34, corresponding to downsampling factors of $\times 16$ and $\times 32$, respectively. Moreover, our implementation of ResNet34 differs from the original one by replacing BatchNorm+ReLU layers with synchronized InPlaceABN (\iABNsync) activated with LeakyReLU with negative slope $0.01$ as proposed in~\cite{RotPorKon18a}. This change allows to free up a significant amount of GPU memory, which can be exploited to scale up the batch size and, therefore, improve the quality of the computed gradients. 
In Fig.~\ref{fig:architecture} we depict our backbone, where white rectangles in the FPN module denote $1\times 1$ or $3\times 3$ convolution layers with $256$ output channels, each followed by \iABNsync.

\paragraph{Inputs.} The backbone takes in input an RGB image $x$. 

\paragraph{Outputs.}
The backbone provides $2$ output tensors, namely $\{f_1, f_2\}$, corresponding to the $2$ different scales of the FPN network with downsampling factors of $\times 16$ and $\times 32$, each with $256$ feature channels (see, Fig.~\ref{fig:architecture}).

\vspace{-10pt}
\subsection{3D Detection Head}\label{ss:head}
We build the 3D detection head by modifying the single-stage 2D detector implemented in RetinaNet~\cite{Lin+17}. 
We apply the detection module independently to each output $f_i$ of our backbone, thus operating at a different scale of the FPN as described above. The detection modules share the same parameters and provide dense 3D bounding boxes predictions. In addition, we let the module regress 2D bounding boxes similar to~\cite{simonelli2019disentangling,brazil2019m3d}, but in contrast to those works, we will not use the predicted 2D bounding boxes but rather consider this as a regularizing side task.
Akin to RetinaNet, this module makes use of so-called \emph{anchors}, which implicitly provide some pre-defined 2D bounding boxes that the network can modify. The number of anchors per spatial location is given by $\con n_a$. 
Fig.~\ref{fig:architecture} shows the architecture of our 3D detection head. It consists of two parallel branches, the top one devoted to providing confidences about the predicted 2D and 3D bounding boxes, while the bottom one is devoted to regressing the actual bounding boxes. White rectangles denote $3\times 3$ convolutions with $128$ output channels followed by \iABNsync. More details about the input and outputs of this module are given below, by following the notation adopted in~\cite{simonelli2019disentangling}.

\paragraph{Inputs.} The 3D detection head takes $f_i$, $i\in\{1,2\}$, \ie~an output tensor of our backbone, as input. Each tensor $f_i$ has a spatial resolution of $w_i\times h_i$.

\paragraph{Outputs.} 
The detection head outputs a 2D bounding box and $\con n_c$ 3D bounding boxes (with confidences) for each anchor $a$ and spatial cell $g$ of the $w_i\times h_i$ grid of $f_i$.
Each anchor $a$ provides a reference size $(w_a,h_a)$ for the 2D bounding box.
The 2D bounding box is given in terms of $\vct\theta_\text{2D}=(\delta_u,\delta_v,\delta_w,\delta_h)$ and $\vct\zeta_\text{2D}=(\zeta_\text{2D}^1, \ldots, \zeta_\text{2D}^{\con n_c})$ from which we can derive
\begin{itemize}
	\item $p^c_\text{2D}=(1+e^{-\zeta^c_\text{2D}})^{-1}$, \ie~the probability that the 2D bbox belongs to class $c$,
        \item $(u_b,v_b)=(u_g+\delta_u w_a, v_g+\delta_v h_a)$, \ie~the bounding box's center, where $(u_g,v_g)$ are the image coordinates of cell $g$, and
	\item $(w_b,h_b)=(w_a e^{\delta_w}, h_a e^{\delta_h})$, \ie~the size of the bounding box.
\end{itemize}
In addition to the 2D bounding box the head returns, for each class $1\leq c\leq \con n_c$, a 3D bounding box in terms of $\vct\theta_\text{3D}=(\Delta_u, \Delta_v, \delta_z, \delta_W, \delta_H, \delta_D, r_x, r_z)$ and $\zeta_\text{3D}$ (we omitted the superscript $c$). Indeed, from those outputs we can compute
\begin{itemize}
\item $p^c_\text{3D$|$2D}=(1+e^{-\zeta_\text{3D}})^{-1}$, \ie~the per-class 3D bbx confidence,
\item $\vct c=(u_b+\Delta_u, v_b+\Delta_v)$, \ie~the 3D bbox center projected on the image plane,
\item $z=\mu^c_z+\sigma^c_z\delta_z$, \ie~the depth of the bounding box center, where $\mu_z^c$ and $\sigma_z^c$ are class- and $\con Z_\text{res}$-specific depth mean and standard deviation,
\item $\vct s=(W^c_0 e^{\delta_W}, H^c_0 e^{\delta_H},D^c_0 e^{\delta_D})$, \ie~the 3D bounding box dimensions, where $(W^c_0, H^c_0, D^c_0)$ is a reference size for 3D bounding boxes belonging to class $c$,
\item $\alpha=\text{atan2}(r_x, r_z)$ is the rotation angle on the $XZ$-plane with respect to an allocentric coordinate system.
\end{itemize}
The actual confidence of each 3D bounding box is computed by combining the 2D and 3D bounding box probabilities into $p_\text{3D}^c=p_\text{3D$|$2D}^cp_\text{2D}^c$.


\paragraph{Losses.}
The losses we employ to regress the 2D bounding boxes and to learn the 2D class-wise confidence are inherited from the RetinaNet 2D detector~\cite{Lin+17}.
Also the logic for the assignment of ground-truth boxes to anchors is taken from the same work, but we use it in a slightly different way, since we have 3D bounding boxes as ground-truth. The idea is to extract the 2D bounding box from the projected 3D bounding box and use this to guide the assignment of the ground-truth box to anchors.
As for the losses pertaining to the 3D detection part, we exploit the lifting transformation combined with the loss disentangling strategy as proposed in~\cite{simonelli2019disentangling}. Indeed, the lifting transformation allows to sidestep the issue of finding a proper way of balancing losses for the different outputs of the network, which inherently operate at different scales, by optimizing a single loss directly at the 3D bounding box level. However, this loss entangles the network's outputs in a way that renders the training dynamics unstable, thus harming the learning process. Nonetheless, this can be overcome by employing the disentangling transformation~\cite{simonelli2019disentangling}. We refer to the latter work for  details.

\vspace{-10pt}
\section{Experiments}\label{sec:exp_general}
In this section we validate our contributions on the KITTI3D dataset \cite{Geiger2012CVPR}. After providing some details about the implementation of our method, we give a description of the dataset and its metrics. Then, we show the results obtained comparing our single-stage architecture \MoVi against state-of-the-art methods on the KITTI3D benchmark. Finally, to better highlight the importance of our novel technical contribution, we perform an in-depth ablation study.

\vspace{-10pt}
\subsection{Implementation details}
In this section we provide the details about the implementation of the virtual views as well as relevant information about the optimization.

\paragraph{Virtual Views.}
We implement our approach using a parametrization that provides good performances without compromising the overall speed of the method. During training we generate a total of $\con n_v = 8$ virtual views per training image, by using a class-uniform, ground-truth-oriented sampling strategy with probability $p_v = 0.7$, random otherwise (see Sec.~\ref{sec:vc}). We set the depth resolution $\con Z_\text{res}$ to $5m$. During inference we limit the search space along depth to $[4.5m, 45m]$. We set the dimensions of all the generated views to have height $\con h_v=100$ pixels and width $\con w_v=331$ pixels. We set the depth statistics as $\mu_z = 3m$ and $\sigma_z = 1m$.

\paragraph{Optimization.}
Our network is optimized in an end-to-end manner and in a \textit{single} training phase, not requiring any multi-step or warm-up procedures. We used SGD with a learning rate set at 0.2 and a weight decay of 0.0001 to all parameters but scale and biases of iABN. Following ~\cite{simonelli2019disentangling}, we did not optimize the parameters in conv1 and conv2 of the ResNet34. Due to the fairly reduced resolution of the virtual views, we are able to train with a batch size of 2048 on 4 NVIDIA V-100 GPUs for 20k iterations, decreasing the learning rate by a factor of 0.1 at 16k and 18k iterations. No form of augmentation (\eg~voting using multi-scale, horizontal flipping, \etc) has been applied during inference.

\paragraph{3D Detection Head.} 
We adopt 18 anchors with six aspect ratios {\{ $\frac{1}{3}$, $\frac{1}{2}$, $\frac{3}{4}$, 1, 2, 3\}} and three different scales \{$2s_i 2^{\frac{j}{3}} : j \in {0, 1, 2}$\}, where $s_i$ is the down-sampling factor of the FPN level $f_i$. Each anchor is considered positive if its IoU with a ground truth object is greater than $0.5$. To account for the presence of objects of different categories and therefore of fairly different 3D extent, we create class-wise \textit{reference anchors}. Each \textit{reference anchor} has been obtained by observing the dataset statistics of the training set. We define the reference \textit{Car} size as $W_0 = 1.63m$, $H_0 = 1.53m$, $D_0 = 3.84m$, the \textit{Pedestrian} reference as $W_0 = 0.63m$, $H_0 = 1.77m$, $D_0 = 0.83m$ and the \textit{Cyclist} reference as $W_0 = 0.57m$, $H_0 = 1.73m$, $D_0 = 1.78m$. 

\paragraph{Losses.}
We used a weight of 1 for the 2D confidence loss and for the 3D regression loss, while we set at 0.5 the weight of the 2D regression and 3D confidence loss. The Huber parameter is set to $\delta_H = 3.0$ and the 3D confidence temperature to $T = 1$ as done in \cite{simonelli2019disentangling}. 

\vspace{-10pt}
\subsection{Dataset and Experimental Protocol}

\paragraph{Dataset.}\label{sec:kitti_descr}
The KITTI3D dataset is arguably the most influential benchmark for monocular 3D object detection. It consists of 7481 training and 7518 test images. Since the dataset does not provide an official validation set, it is common practice to split the training data into 3712 training and 3769 validation images as proposed in~\cite{Chen2015} and then report validation results. For this reason it is also mandatory not to limit the analysis of the results to the validation set but instead to provide results on the official test set obtained via the KITTI3D benchmark evaluation server\footnote{Official KITTI3D benchmark: \url{www.cvlibs.net/datasets/kitti/eval\_object.php?obj\_benchmark=3d}\label{foot:kitti_page}}. The dataset annotations are provided in terms of 3D bounding boxes, each one characterized by a \textit{category} and a \textit{difficulty}. The possible object categories are \textit{Car}, \textit{Pedestrian} and \textit{Cyclist}, while the object difficulties are chosen among \textit{Easy}, \textit{Moderate} and \textit{Hard} depending on the object distance, occlusion and truncation. It is also relevant to note that the number of per-class annotations is profoundly different, causing the dataset to have a fairly high class imbalance. On a total of 28,8k annotations, 23.0k (79.8\%) are \textit{Car} objects, while 4.3k (15.0\%) are \textit{Pedestrian} and only 1.5k (5.2\%) are \textit{Cyclist}.

\paragraph{Experimental Protocol.}\label{sec:exp_proto}
In order to provide a fair comparison, we followed the experimental protocol of M3D-RPN \cite{brazil2019m3d} and SS3D \cite{jorgensed2019ss3d}, \ie the only other available multi-class, monocular, RGB-only methods. To this end, we show results on all the KITTI3D classes obtained by means of a \textit{single multi-class} model. For completeness we also report results of other methods (\eg~single-class or RBG+LiDAR), but we remark that a fair comparison is only possible with ~\cite{brazil2019m3d,jorgensed2019ss3d}.

\paragraph{Evaluation Protocol.}\label{sec:eval_proto}
Our results follow the Official KITTI3D protocol~\footnoteref{foot:kitti_page}. In particular, we report scores in terms of the official 3D Average Precision (AP) metric and Bird's Eye View (BEV) AP metric. These scores have been computed with the official class-specific thresholds which are 0.7 for \textit{Car} and 0.5 for \textit{Pedestrian} and \textit{Cyclist}.
Recently, there has been a modification in the KITTI3D metric computation. The previous $AP|_{R_{11}}$ metric, which has been demonstrated to provide biased comparisons~\cite{simonelli2019disentangling}, has been deprecated in favour of the $AP|_{R_{40}}$. Due to this fact, we invite to refer only to $AP|_{R_{40}}$ and to disregard any score computed with $AP|_{R_{11}}$. Even if a method was published before the new metric has been introduced, its updated $AP|_{R_{40}}$ test scores should be visible online~\footnoteref{foot:kitti_page}. The availability of validation scores is instead dependent on the publication date and to the author's willingness to provide results with the new unbiased metric.

\begin{table*}[t!]
    \centering
    \caption{Test set SOTA results on \textit{Car} (0.7 IoU threshold)}
    \vspace{5pt}
    \resizebox{.8\columnwidth}{!}{
    {\footnotesize
    \begin{tabular}{l|c|c|ccc|ccc}    
        \toprule
         & \# & Training &\multicolumn{3}{c|}{3D detection} & \multicolumn{3}{c}{Bird's eye view} \\
        Method & classes & data & Easy & Moderate & Hard & Easy & Moderate & Hard \\
        \midrule
        OFTNet~\cite{Roddick18}           & single & RGB 	        &  1.61 &  1.32 &  1.00 	&  7.16  &  5.69 &  4.61 \\
        FQNet~\cite{Liu+19}               & single & RGB            &  2.77 &  1.51 &  1.01 	&  5.40  &  3.23 &  2.46 \\
        ROI-10D~\cite{Manhardt_2019_CVPR} & single & RGB+Depth 	    &  4.32 &  2.02 &  1.46 	&  9.78  &  4.91 &  3.74 \\
        GS3D~\cite{li2019gs3d}            & single & RGB            &  4.47 &  2.90 &  2.47 	&  8.41  &  6.08 &  4.94 \\
        MonoGRNet~\cite{qin2019monogrnet} & single & RGB 	        &  9.61 &  5.74 &  4.25 	& 18.19  & 11.17 &  8.73 \\
        \monodis{}~\cite{simonelli2019disentangling} & single & RGB & 10.37 &  7.94 &  6.40 	& 17.23  & 13.19 & 11.12 \\
        MonoPSR~\cite{ku2019monopsr}      & single & RGB+LiDAR      & 10.76 &  7.25 &  5.85 	& 18.33  & 12.58 &  9.91 \\
        SS3D~\cite{jorgensed2019ss3d}     & multi  & RGB            & 10.78 &  7.68 &  6.51  	& 16.33  & 11.52 &  9.93 \\
        SMOKE~\cite{liu2020smoke}         & single & RGB             & 14.03 &  9.76 &  7.84     & 20.83  & 14.49 & 12.75 \\
        M3D-RPN~\cite{brazil2019m3d}      & multi & RGB 	        & 14.76 &  9.71 &  7.42	    & 21.02  & 13.67 & 10.23 \\
        \rowcolor{mapillarygreen}
        Ours  & multi & RGB & \textbf{15.19} & \textbf{10.90} & \textbf{9.26} & \textbf{22.76} & \textbf{17.03} & \textbf{14.85} \\
        \bottomrule
    \end{tabular}}}
    \label{tab:kitti_test_car}
\end{table*}
\begin{table*}[t!]
    \centering
    \caption{Test set SOTA results on \textit{Pedestrian} and \textit{Cyclist} (0.5 IoU threshold)}
    \vspace{5pt}
    \resizebox{1.0\columnwidth}{!}{
    {\footnotesize
    \begin{tabular}{l|c|c|ccc|ccc|ccc|ccc}    
        \toprule
         & & &\multicolumn{6}{c|}{Pedestrian} & \multicolumn{6}{c}{Cyclist} \\
         & \# & Training &\multicolumn{3}{c}{3D Detection} & \multicolumn{3}{c|}{Bird's eye view} & \multicolumn{3}{c|}{3D Detection} & \multicolumn{3}{c}{Bird's eye view} \\
        Method & classes & data & Easy & Moderate & Hard & Easy & Moderate & Hard & Easy & Moderate & Hard & Easy & Moderate & Hard \\
        \midrule
        OFTNet~\cite{Roddick18} & single & RGB                   & 0.63 & 0.36 & 0.35   & 1.28 & 0.81 & 0.51   & 0.14 & 0.06 & 0.07 & 0.36 & 0.16 & 0.15 \\
        SS3D~\cite{jorgensed2019ss3d} & multi & RGB             & 2.31 & 1.78 & 1.48   & 2.48 & 2.09 & 1.61   & 2.80 & 1.45 & 1.35 & 3.45 & 1.89 & 1.44 \\
        M3D-RPN~\cite{brazil2019m3d} & multi & RGB               & 4.92 & 3.48 & 2.94   & 5.65 & 4.05 & 3.29   & 0.94 & 0.65 & 0.47 & 1.25 & 0.81 & 0.78\\
        MonoPSR~\cite{ku2019monopsr} & single & RGB+LiDAR        & 6.12 & 4.00 & 3.30   & 7.24 & 4.56 & 4.11   &  \textbf{8.37} &  \textbf{4.74} &  \textbf{3.68} &  \textbf{9.87} &  \textbf{5.78} &  \textbf{4.57} \\
        \rowcolor{mapillarygreen}
        Ours  & multi & RGB                                      &  \textbf{8.99} &  \textbf{5.44} &  \textbf{4.57} &  \textbf{10.08} &  \textbf{6.29} &  \textbf{5.37} & 1.08 & 0.63 & 0.70 & 1.45 & 0.91 & 0.93\\
        \bottomrule
    \end{tabular}}}
    \label{tab:kitti_test_ped_cyc}
    \vspace{-10pt}
\end{table*}

\begin{table*}[t]
    \centering
    \vspace{-10pt}
    \caption{Validation set results on all KITTI3D classes. (0.7 IoU threshold on \textit{Car}, 0.5 on \textit{Pedestrian} and \textit{Cyclist}). \textbf{V} = Virtual Views, \textbf{B} = Bin-based estimation}
    \vspace{5pt}
    \resizebox{1.0\columnwidth}{!}{
    {\footnotesize
    \begin{tabular}{c|c|ccc|ccc|ccc|ccc|ccc|ccc}    
        \toprule
         &       &\multicolumn{6}{c|}{Car} &\multicolumn{6}{c|}{Pedestrian} & \multicolumn{6}{c}{Cyclist} \\
         &  &\multicolumn{3}{c}{3D Detection} & \multicolumn{3}{c|}{Bird's eye view} & \multicolumn{3}{c|}{3D Detection} & \multicolumn{3}{c}{Bird's eye view}  & \multicolumn{3}{c|}{3D Detection} & \multicolumn{3}{c}{Bird's eye view}\\
        Method & $\con Z_\text{res}$  & Easy & Mod. & Hard & Easy & Mod. & Hard & Easy & Mod. & Hard & Easy & Mod. & Hard & Easy & Mod. & Hard & Easy & Mod. & Hard \\
        \midrule
        \midrule
        MonoDIS~\cite{simonelli2019disentangling} & --     & 11.06 &  7.60 &  6.37 & 18.45 & 12.58 & 10.66 & 3.20 & 2.28 & 1.71 & 4.04 & 3.19 & 2.45 & 1.52 & 0.73 & 0.71 & 1.87 & 1.00 & 0.94 \\
        \midrule
        MonoDIS+\textbf{V}  & 5m     & \textbf{13.40} & \textbf{10.89} &  \textbf{9.67} & \textbf{21.90} & \textbf{17.38} & \textbf{15.71} & \textbf{4.98} & \textbf{3.31} & \textbf{3.06} & \textbf{6.83} & \textbf{4.33} & \underline{3.38} & \underline{2.09} & \underline{1.07} & \underline{1.00} & \underline{2.70} & \underline{1.42} & \underline{1.31} \\
        MonoDIS+\textbf{B}  & 5m     & 7.30  &  5.34 &  4.25 & 12.83 &  8.77 &  7.21 & \underline{3.96} & 3.10 & 2.49 & \underline{4.87} & \underline{3.65} & 3.01 & 0.44 & 0.31 & 0.26 & 0.82 & 0.39 & 0.27 \\
        MonoDIS+\textbf{B}  &10m     & \underline{11.64} &  \underline{8.36} &  \underline{7.25} & \underline{19.07} & \underline{12.98} & \underline{11.39} & 3.37 & \underline{3.13} & \underline{2.53} & 4.56 & 4.21 & \textbf{3.44} & \textbf{2.76} & \textbf{1.80} & \textbf{1.72} & \textbf{3.39} & \textbf{2.20} & \textbf{2.18} \\
        \midrule
        \midrule
        \MoVi  & 5m     & \textbf{14.28} & \textbf{11.13} &  \textbf{9.68} & \textbf{22.36} & \textbf{17.87} & \textbf{15.73} & \textbf{7.86} & \textbf{5.52 }& \textbf{4.42} & \textbf{9.25} & \textbf{6.63} & \textbf{5.06 }& \textbf{2.63} & \textbf{1.27} & \textbf{1.13} & \textbf{3.10} & \textbf{1.57} & \textbf{1.30} \\
        \MoVi          & 10m    & \underline{11.58} &  \underline{9.54} &  \underline{8.54} & \underline{17.98} & \underline{15.16} & \underline{13.98} & \underline{1.82} & \underline{1.27} & \underline{0.94} & \underline{2.38} & \underline{1.78} & \underline{1.34} & \underline{1.08} & \underline{0.51} & \underline{0.51} & \underline{1.84} & \underline{0.97} & \underline{0.89} \\
        \MoVi         & 20m    &  7.68 &  6.18 &  5.56 & 13.35 & 11.11 & 10.22 & 1.55 & 0.97 & 0.83 & 1.97 & 1.39 & 1.05 & 0.25 & 0.10 & 0.10 & 0.36 & 0.17 & 0.17 \\
        \bottomrule
    \end{tabular}}}
    \label{tab:kitti_ablations_bands}
\end{table*}
\begin{table*}[t]
    \centering
    \vspace{-10pt}
        \caption{Ablation results on \textit{Car} obtained on different distance ranges.}
        \vspace{5pt}
        \resizebox{0.5\columnwidth}{!}{
        \begin{tabular}{l|c|c|ccc|ccc}
            \toprule
            & train & val &\multicolumn{3}{c|}{3D detection} & \multicolumn{3}{c}{Bird's eye view} \\
            Method & range & range & Easy & Mod. & Hard & Easy & Mod. & Hard \\
            \midrule
            \midrule
            MonoDIS~\cite{simonelli2019disentangling} & far & near         &  0.2 &  0.1 & 0.1 &  0.2 &  0.1 &  0.1 \\
            \MoVi                                   & far & near          &  \textbf{4.0} &  \textbf{1.9} & \textbf{1.7} &  \textbf{5.5} &  \textbf{2.7} &  \textbf{2.4} \\
            \midrule
            MonoDIS~\cite{simonelli2019disentangling} & near & far         &  0.2 &  0.1 & 0.1 &  0.2 &  0.2 &  0.1 \\
            \MoVi                                   & near & far          &  \textbf{3.3} &  \textbf{1.4} & \textbf{1.7} &  \textbf{4.2} &  \textbf{1.9} &  \textbf{2.3} \\
            \midrule
            MonoDIS~\cite{simonelli2019disentangling} & near+far & middle  &  0.6 &  0.4 & 0.3 &  0.7 &  0.5 &  0.4 \\
            \MoVi                                   & near+far & middle   & \textbf{19.2} & \textbf{10.6} & \textbf{8.8 }& \textbf{22.9} & \textbf{12.8} & \textbf{10.4} \\
            \bottomrule
        \end{tabular}}
        \label{tab:kitti_ablation_distance}
    \vspace{-10pt}
\end{table*}

\vspace{-10pt}
\subsection{3D Detection}
In this section we show the results of our approach, providing a comparison with state-of-the-art 3D object detection methods. As previously stated in Sec.~\ref{sec:exp_proto}, we would like to remind that some of the reported methods do not adopt the same experimental protocol as ours. Furthermore, due to the formerly mentioned redefinition of the metric computation, the performances reported by some previous methods which used a potentially biased metric cannot be taken into consideration. For this reason we focus our attention on the performances on the \textit{test} split, reporting official results computed with the updated metric.

\paragraph{Performances on class Car.}
In Tab.~\ref{tab:kitti_test_car} we show the results on class \textit{Car} of the KITTI3D test set. It is evident that our approach outperforms all baselines on both 3D and BEV metrics, often by a large margin. In particular, our method achieves better performances compared to \textit{single class} models (\eg~MonoDIS~\cite{simonelli2019disentangling}, MonoGRNet~\cite{qin2019monogrnet}, SMOKE~\cite{liu2020smoke}) and to methods which use LiDAR information during training (MonoPSR~\cite{ku2019monopsr}). Our method also outperforms the other \textit{single-stage}, \textit{multi-class} competitors (M3D-RPN~\cite{brazil2019m3d}, SS3D~\cite{jorgensed2019ss3d}). This is especially remarkable considering the fact that M3D-RPN relies on a fairly deeper backbone (DenseNet-121) and, similarly to SS3D, it also uses a post-optimization process and a multi-stage training. It is also worth noting that our method achieves the largest improvements on \textit{Moderate} and \textit{Hard} sets where object are in general more distant and occluded: on the 3D AP metric we improve with respect to the best competing method by \textbf{+12.3\%} and \textbf{+24.8\%} respectively while for the BEV AP metric improves by \textbf{+24.6\%} and \textbf{+33.5\%}, respectively. 

\paragraph{Performances on the other KITTI3D classes.}
In Tab.~\ref{tab:kitti_test_ped_cyc} we report the performances obtained on the classes \textit{Pedestrian} and \textit{Cyclist} on the KITTI3D test set. On the class \textit{Pedestrian} our approach outperforms all the competing methods on all levels of difficulty considering both 3D AP and BEV AP. Remarkably, we also achieve better performance than MonoPSR~\cite{ku2019monopsr} which exploits LiDAR at training time, in addition to RGB images. The proposed method also outperforms the \textit{multi-class} models in~\cite{brazil2019m3d,jorgensed2019ss3d}. On \textit{Cyclist} our method achieves modest improvements with respect to M3D-RPN~\cite{brazil2019m3d}, but it does not achieve better performances than SS3D~\cite{jorgensed2019ss3d} and MonoPSR~\cite{ku2019monopsr}. However, we would like to remark that MonoPSR~\cite{ku2019monopsr} exploits additional source of information (\ie~ LiDAR) besides RGB images, while SS3D~\cite{jorgensed2019ss3d} underperforms on \textit{Car} and \textit{Pedestrian} which, as described in Sec.~\ref{sec:kitti_descr}, are the two most represented classes.

\paragraph{Ablation studies.}
In Tab.~\ref{tab:kitti_ablations_bands},\ref{tab:kitti_ablation_distance} we provide three different ablation studies. 
First, in $1^{st}$-$4^{th}$ row of Tab.~\ref{tab:kitti_ablations_bands} we put our proposed virtual views in comparison with a bin-based distance estimation approach. To do so, we took a common baseline, MonoDIS~\cite{simonelli2019disentangling}, and modified it in order to work with both virtual views and bin-based estimation. The $1^{st}$ row of Tab.~\ref{tab:kitti_ablations_bands} shows the baseline results of MonoDIS as reported in~\cite{simonelli2019disentangling}. In the $2^{nd}$ row we report the scores obtained by applying our virtual views to MonoDIS (MonoDIS+\textbf{V}). In the $3^{rd}$-$4^{th}$ rows we show the results obtained with MonoDIS with the bin-based approach (MonoDIS+\textbf{B}). For these last experiments we kept the full-resolution image as input, divided the distance range into $\con Z_{\text{res}}$-spaced bins, assigned each object to a specific bin, learned this assignment as a classification task and finally applied a regression-based refinement. By experimenting with different $\con Z_{\text{res}}$ values, we found that MonoDIS+\textbf{V} performs best with  $\con Z_{\text{res}}=5m$ while MonoDIS+\textbf{B} performed best with $\con Z_{\text{res}}=10m$. With the only exception of the class \textit{Cyclist}, the MonoDIS+\textbf{V} outperforms MonoDIS+\textbf{B} in both 3D and BEV AP. 
Second, in the $3^{rd}$-$6^{th}$ rows of Tab.~\ref{tab:kitti_ablations_bands} we show the results of another ablation study in which we focus on different possible $\con Z_{\text{res}}$ configurations of our proposed \MoVi detector. In this regard, we show the performances by setting $\con Z_\text{res}$ to $5m$ ($3^{rd}$ row), $10m$ ($4^{th}$) and $20m$ ($5^{th}$). Among the different settings, the depth resolution $\con Z_\text{res}=5m$ outperforms the others by a clear margin. 
Finally, in Tab.~\ref{tab:kitti_ablation_distance} we conduct another ablation experiment in order to measure the generalization capabilities of our virtual views. We create different versions of the KITTI3D train/val splits, each one of them containing objects included into a specific \textit{depth range}. In particular, we define a \textit{far/near} train/val split, where the depth of the objects in the training split is in [0m, 20m] whereas the depth of the objects included into the validation split is in [20m, 50m]. We then define a \textit{near/far} train/val split by reversing the previous splits, as well as a third train/val split regarded as \textit{near+far/middle} where the training split includes object with depth in [0m,10m] + [20m, 40m] while the validation is in [10m, 20m]. We compare the results on these three train/val splits with the MonoDIS~\cite{simonelli2019disentangling} baseline, decreasing the AP IoU threshold to 0.5 in order to better comprehend the analysis. By analyzing the results in Tab.~\ref{tab:kitti_ablation_distance} it is clear that our method generalizes better across ranges, achieving performances which are one order of magnitude superior to the baseline.    

\paragraph{Inference Time.} Inference time plays a key role for the application of monocular 3D object detection methods. Our method demonstrates to achieve real-time performances reaching, under the best configuration with $\con Z_{\text{res}} = 5m$, an average inference time of 45ms. As expected, the inference time is influenced by the number of views. We found the inference time to be inversely proportional to the discretization of the distance range $\con Z_{\text{res}}$. In fact, we observe that inference time goes from 13ms with $\con Z_{\text{res}}=20m$, to 25ms (10m), 45ms (5m).

\begin{figure*}[t]
    \centering
    \includegraphics[width=0.8\textwidth]{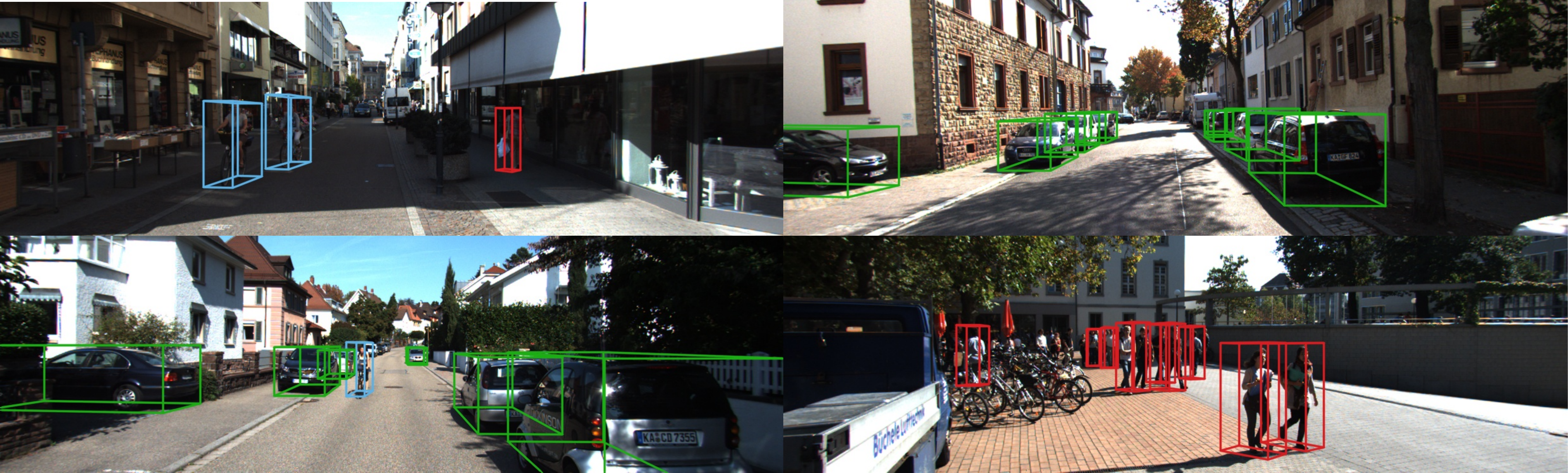}
    \caption{Qualitative results obtained with \MoVi on KITTI3D.}
    \label{fig:qualitative}
    \vspace{-10pt}
\end{figure*}

\begin{figure*}[h]
    \centering
    \includegraphics[width=.95\textwidth]{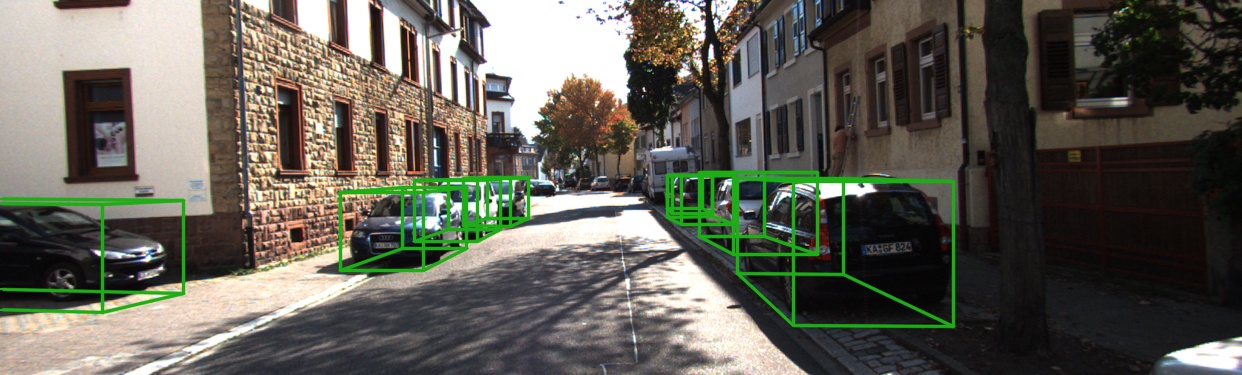}
    \includegraphics[width=.95\textwidth]{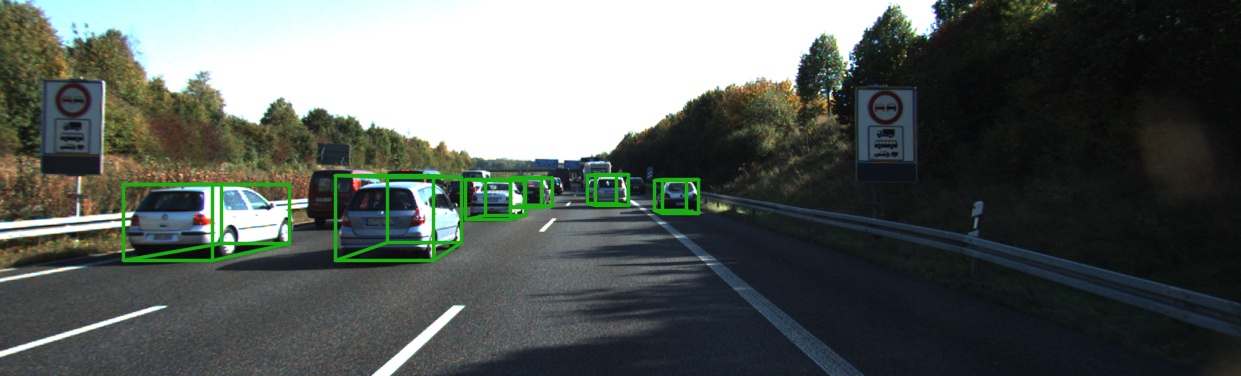}
    \includegraphics[width=.95\textwidth]{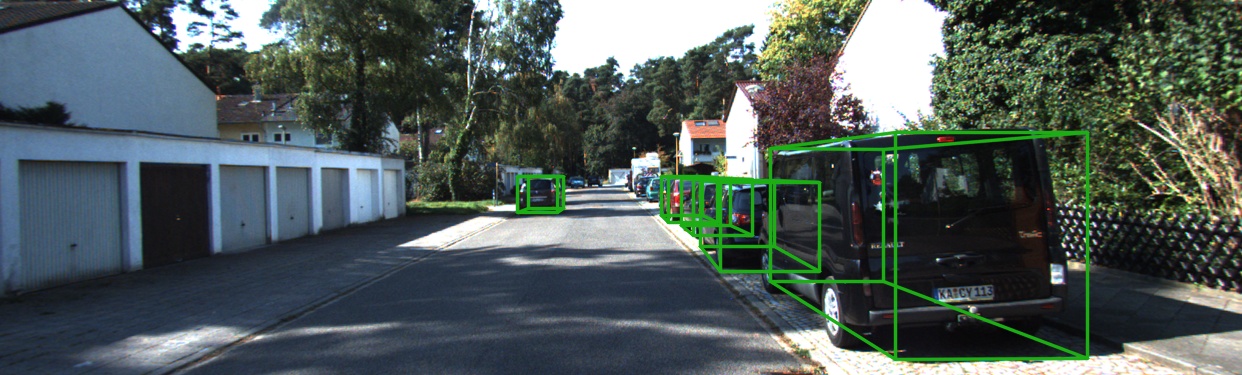}
    \includegraphics[width=.95\textwidth]{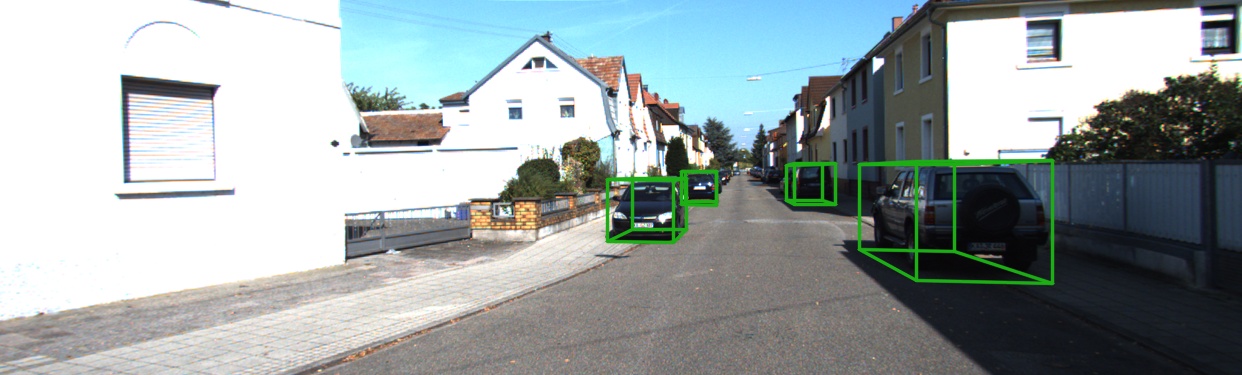}
    \includegraphics[width=.95\textwidth]{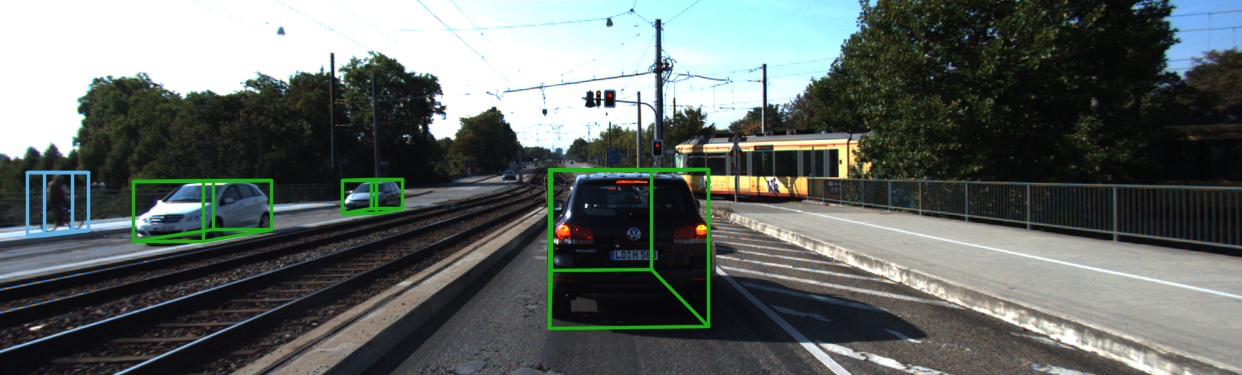}
    \caption{Example results of our MoVi-3D model on KITTI3D validation images.}
    \label{fig:supp_1}
\end{figure*}

\begin{figure*}[h]
    \centering
    \includegraphics[width=.95\textwidth]{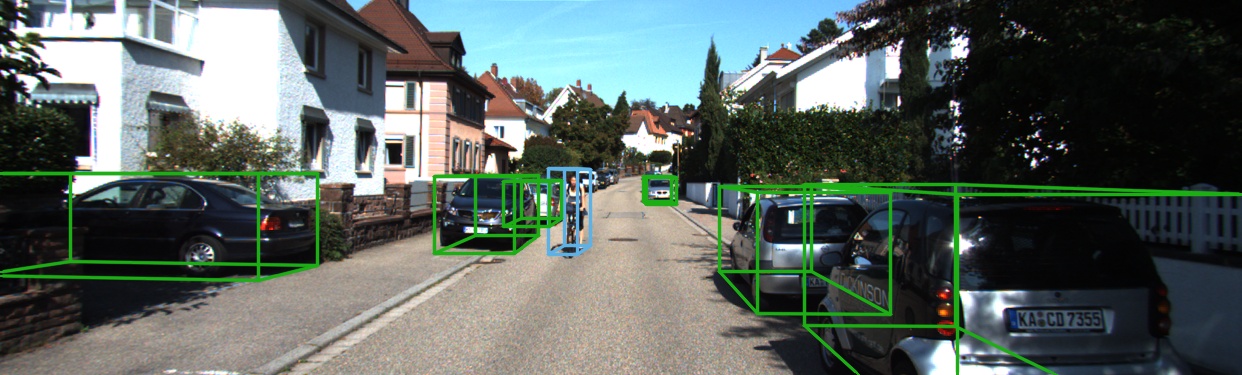}
    \includegraphics[width=.95\textwidth]{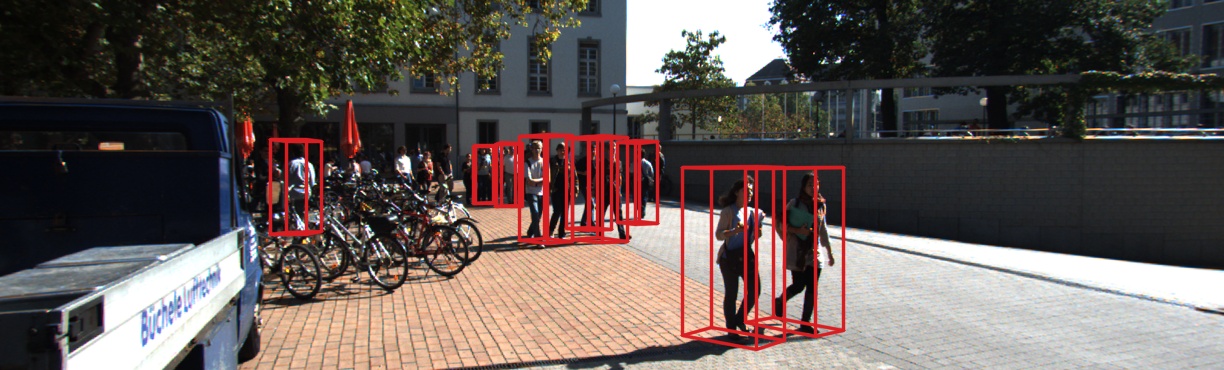}
    \includegraphics[width=.95\textwidth]{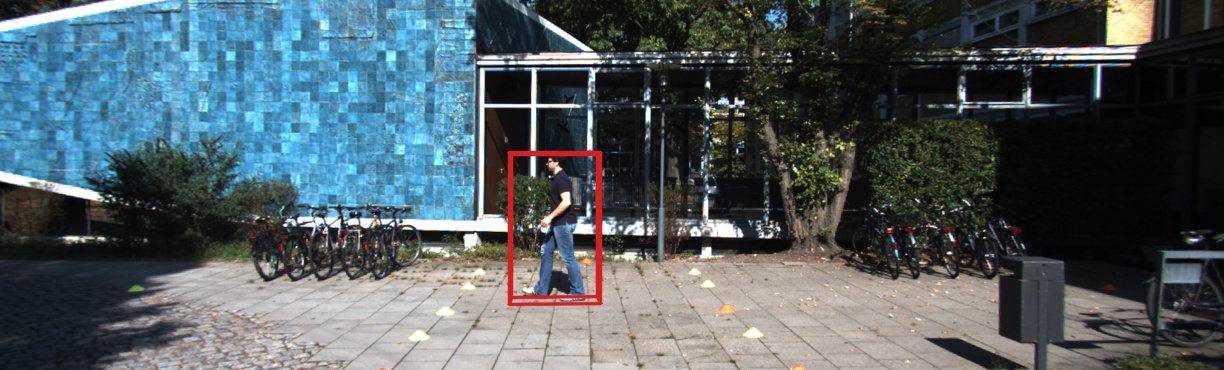}
    \includegraphics[width=.95\textwidth]{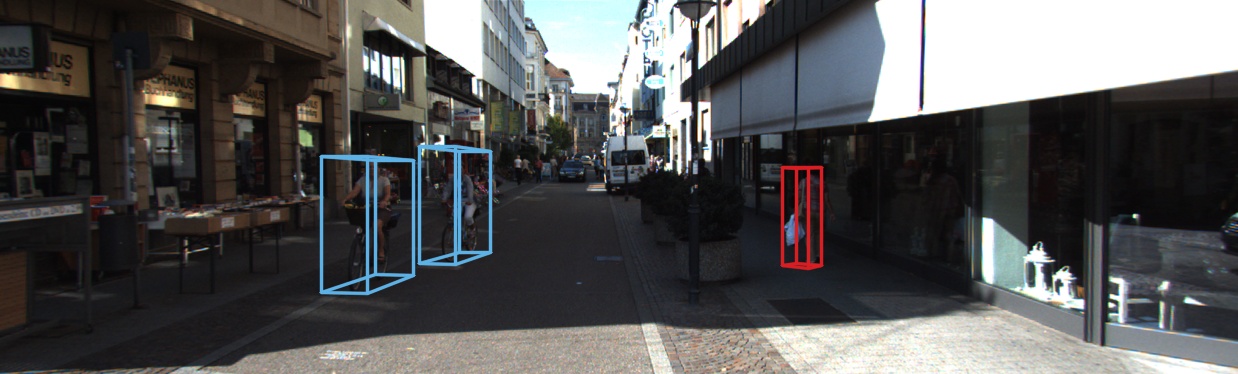}
    \includegraphics[width=.95\textwidth]{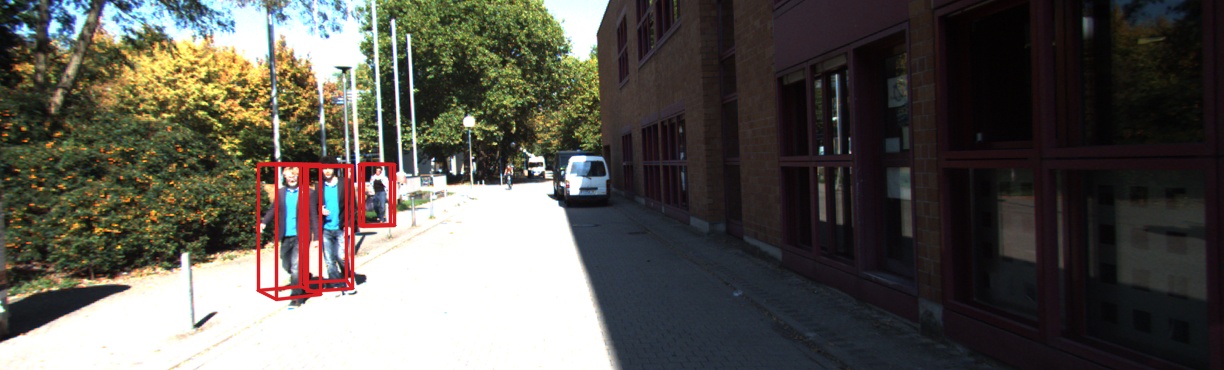}
    \caption{Further example results of our MoVi-3D model on KITTI3D validation images.}
    \label{fig:supp_2}
\end{figure*}

\paragraph{Qualitative results.} We provide some qualitative results in Fig.~\ref{fig:qualitative}. We also provide full-size qualitative results in Fig.~\ref{fig:supp_1},\ref{fig:supp_2}.

\vspace{-10pt}
\section{Conclusions}

We introduced new training and inference schemes for 3D object detection from single RGB images, designed with the purpose of injecting depth invariance into the model. 
At training time, our method generates virtual views that are positioned within a small neighborhood of the objects to be detected. This yields to learn a model that is supposed to detect objects within a small depth range independently from where the object was originally positioned in the scene. 
At inference time, we apply the trained model to multiple virtual views that span the entire range of depths at a resolution that relates to the depth tolerance considered at training time.
Due to the gained depth invariance, we also designed a novel, lightweight, single-stage deep architecture for 3D object detector that does not make explicit use of regressed 2D bounding boxes at inference time, as opposite to many previous methods. Overall, our approach achieves state-of-the-art results on the KITTI3D benchmark. 
Future research will focus on devising data-driven methods to adaptively generate the best views 
at inference time. 

\vspace{-10pt}
\bibliographystyle{splncs04}

\input{arxiv.bbl}
\end{document}

%% file: intro.tex
With the advent of autonomous driving, significant attention has been devoted in the computer vision and robotics communities to the semantic understanding of urban scenes. In particular, object detection is one of the most prominent challenges that must be addressed in order to build autonomous vehicles able to drive safely over long distances. 
In the last decade, thanks to the emergence of deep neural networks and to the availability of large-scale annotated datasets, the state of the art in 2D object detection has improved significantly~\cite{Ren+15,Liu2016,Redmon2016,Redmon_2017_CVPR,Lin+17,Law_2018_ECCV}, reaching near-human performance~\cite{Liu_2018_DetSurvey}.
However, detecting objects in the image plane and, in general, reasoning in 2D, is not sufficient for autonomous driving applications. Safe navigation of self-driving cars requires accurate 3D localization of vehicles, pedestrians and, in general, any object in the scene. 
 As a consequence, depth information is needed. While depth can be obtained from expensive LiDAR sensors or stereo camera rigs, recently, there has been an increasing interest in replacing them with cheaper sensors, such as RGB cameras. Unsurprisingly, state-of-the-art 3D detection methods exploit a multi-modal approach, combining data from RGB images with LiDAR information~\cite{Liang_CVPR_2019,Wang_arxiv_2019,Shin_arxiv_18,shi2018pointrcnn}. However, recent works have  attempted to recover the 3D location and pose of objects from a monocular RGB input \cite{simonelli2019disentangling,brazil2019m3d,ku2019monopsr}, with the ultimate goal of replacing LiDAR with cheaper sensors such as off-the-shelf cameras. Despite the ill-posed nature of the problem, these works have shown that it is possible to infer the 3D position and pose of vehicles in road scenes given a single image with a reasonable degree of accuracy.

This work advances the state of the art by introducing \MoVi, a novel, \textit{single-stage} architecture for \textit{\textbf{Mo}nocular \textbf{3D} object detection}, and new training and inference schemes, which enable the possibility for the model to generalize across depth by exploiting \emph{\textbf{Vi}rtual views}. A virtual view is an image transformation that uses geometrical prior knowledge to factor out the variability in the scale of objects due to depth. Each transformation is related to a predefined 3D viewport in space with some prefixed size in meters, \ie a 2D window parallel to the image plane that is ideally positioned in front of an object to be detected, and provides a virtual image of the scene visible through the viewport from the original camera, re-scaled to fit a predetermined resolution (see Fig.~\ref{fig:teaser}). By doing so, no matter the depth of the object, its appearance in the virtual image will be consistent in terms of scale. This allows to partially sidestep the burden of learning depth-specific features that are needed to distinguish objects at different depths, thus enabling the use of simpler models. Also we can limit the range of depths where the network is supposed to detect objects, because we will make use of multiple 3D viewports both at training and inference time.
For this reason, we can tackle successfully the 3D object detection problem with a lightweight, single-stage architecture in the more challenging multi-class setting.

We evaluate the proposed virtual view generation procedure in combination with our \textit{single-stage} architecture on the KITTI 3D Object Detection benchmark \cite{Geiger2012CVPR}, comparing with state-of-the-art methods, and perform an extensive ablation study to assess the validity of our architectural choices. Thanks to our novel training strategy, despite its simplicity, our method is currently the best performing monocular 3D object detection method on KITTI3D that makes no use of additional information at both training and inference time. 

%% file: related.tex
We review recent works on monocular 3D object detection covering both approaches based on RGB-only and those using depth, pseudo-LiDAR or 3D shape information. 

\paragraph{RGB data only.} 
3D object detection from RGB input is inherently challenging as depth information is not available. To compensate for the ill-posed nature of the problem, previous approaches have devised different strategies.
Deep3DBox~\cite{Mousavian_2017_CVPR} proposes to estimate the full 3D pose and the dimensions of objects from 2D bounding boxes by considering projective geometry constraints. 
OFTNet~\cite{Roddick18} considers an orthographic feature transform to map image-level features into a 3D voxel map. The map is subsequently reduced to a 2D representation (birds-eye view). 
Mono3D~\cite{Chen_2016_CVPR} generates 3D candidate boxes and scores them according to semantic segmentation, contextual, shape and class-specific features. At test time boxes are computed based on RGB images only, but the method requires semantic and instance segmentation maps as input.

ROI-10D~\cite{Manhardt_2019_CVPR} proposes a deep architecture equipped with a loss that operates directly in the space of 3D bounding boxes.
In~\cite{Liu+19}, the authors employ a deep network to compute the fitting degree between the proposals and the object in terms of 3D IoU scores, and introduce an approach to filter the estimated box proposals based on 2D object cues only. 
To 
avoid computing features only from 2D proposals, 
GS3D~\cite{li2019gs3d} proposes to derive a coarse cuboid and to extract features from the visible surfaces of the projected cuboid.

MonoGRNet~\cite{qin2019monogrnet} uses a deep network with four specialized modules for different tasks: 2D detection, instance depth estimation, 3D location estimation and local corner regression. The network operates by first
computing the depth and 2D projection of the 3D box center and then estimating the corner coordinates locally.
MonoDIS~\cite{simonelli2019disentangling} describes a two-stage architecture for monocular 3D object detection which disentangles dependencies of different parameters by introducing a novel loss enabling to handle groups of parameters separately. 
MonoPSR~\cite{ku2019monopsr} uses a deep network to jointly compute 3D bounding boxes from 2D ones and estimate 
instance point clouds in order to recover shape and scale information.
SMOKE~\cite{liu2020smoke} proposes to solve the detection task by means of a pixel-based regression and key-point estimation. The regression is performed utilizing a variation of the disentangled loss introduced in \cite{simonelli2019disentangling}.

M3D-RPN~\cite{brazil2019m3d} and SS3D~\cite{jorgensed2019ss3d} are the most closely-related approaches to ours. 
They also implement a single-stage multi-class model. In particular, the former proposes an end-to-end region proposal network using canonical and depth-aware convolutions to generate the predictions, which are then fed to a post-optimization module. SS3D~\cite{jorgensed2019ss3d} proposes to detect 2D key-points as well as predict object characteristics with their corresponding uncertainties. Similarly to M3D-RPN, the predictions are subsequently fed to an optimization procedure to obtain the final predictions. 
Both M3D-RPN and SS3D apply a post-optimization phase 
and, differently from our approach, these methods benefit from a multi-stage training procedure. 

\paragraph{Including depth or pseudo-LiDAR.} 
Some works are based on the idea that more accurate 3D detections can be obtained with the support of depth maps or pseudo-LiDAR point clouds automatically generated from image input. For instance, ROI-10D~\cite{Manhardt_2019_CVPR} exploits depth maps inferred with SuperDepth~\cite{Pillai_2019_ICRA}. 
Pseudo-Lidar~\cite{wang2019pseudo} takes advantage of pre-computed depth maps to convert RGB images to 3D point clouds. Then, state-of-the-art LiDAR-based 3D object detection methods are employed. Pseudo-Lidar++~\cite{yang2019pseudopp} improves over the Pseudo-LiDAR framework adapting the stereo network architecture and the loss function for direct depth estimation, thus producing more accurate predictions for far away objects. 

\paragraph{Including 3D shape information.}
3D-RCNN~\cite{Kundu_2018_CVPR} proposes a convolutional network based on inverse-graphics which maps image regions to the 3D shape and pose of an object instance.
In~\cite{Zia_2014_CVPR} the problem of scene understanding is addressed from the perspective of
3D shape modeling, and a 3D scene representation
is proposed to jointly reason about the 3D shape of multiple objects.
Deep-MANTA~\cite{Chabot_2017_CVPR} uses a multi-task deep architecture for simultaneous vehicle detection, part localization, part visibility characterization and 3D dimension estimation. 
Mono3D++~\cite{TongHe_2019_arxiv} uses a morphable wireframe model for estimating vehicles' 3D shape and pose and optimizes a projection consistency loss between the generated 3D hypotheses and the corresponding 2D pseudo-measurements.
In~\cite{Murthy_17_ICRA} a shape-aware scheme is proposed in order to estimate the 3D pose and shape of a vehicle, given the shape priors encoded in form of keypoints. 

%% file: arxiv.bbl
\begin{thebibliography}{10}
\providecommand{\url}[1]{\texttt{#1}}
\providecommand{\urlprefix}{URL }
\providecommand{\doi}[1]{https://doi.org/#1}

\bibitem{brazil2019m3d}
Brazil, G., Liu, X.: {M3D-RPN}: Monocular 3d region proposal network for object
  detection. In: ICCV. pp. 9287--9296 (2019)

\bibitem{Chabot_2017_CVPR}
Chabot, F., Chaouch, M., Rabarisoa, J., Teuliere, C., Chateau, T.: Deep manta:
  A coarse-to-fine many-task network for joint 2d and 3d vehicle analysis from
  monocular image. In: CVPR (July 2017)

\bibitem{Chen2015}
Chen, T., Li, M., Li, Y., Lin, M., Wang, N., Wang, M., Xiao, T., Xu, B., Zhang,
  C., Zhang, Z.: Mxnet: {A} flexible and efficient machine learning library for
  heterogeneous distributed systems. CoRR  \textbf{abs/1512.01274} (2015)

\bibitem{Chen_2016_CVPR}
Chen, X., Kundu, K., Zhang, Z., Ma, H., Fidler, S., Urtasun, R.: Monocular 3d
  object detection for autonomous driving. In: CVPR (2016)

\bibitem{Geiger2012CVPR}
Geiger, A., Lenz, P., Urtasun, R.: Are we ready for autonomous driving? {T}he
  kitti vision benchmark suite. In: CVPR (2012)

\bibitem{He2015b}
He, K., Zhang, X., Ren, S., Sun, J.: Deep residual learning for image
  recognition. CoRR  \textbf{abs/1512.03385} (2015)

\bibitem{TongHe_2019_arxiv}
He, T., Soatto, S.: Mono3d++: Monocular 3d vehicle detection with two-scale 3d
  hypotheses and task priors. CoRR  \textbf{abs/1901.03446} (2019)

\bibitem{jorgensed2019ss3d}
Jorgensen, E., Zach, C., Kahl, F.: Monocular 3d object detection and box
  fitting trained end-to-end using intersection-over-union loss. In: CVPR
  (2019)

\bibitem{ku2019monopsr}
Ku, J., Pon, A.D., Waslander, S.L.: Monocular 3d object detection leveraging
  accurate proposals and shape reconstruction. In: CVPR (2019)

\bibitem{Kundu_2018_CVPR}
Kundu, A., Li, Y., Rehg, J.M.: {3D-RCNN}: Instance-level 3d object
  reconstruction via render-and-compare. In: (CVPR) (June 2018)

\bibitem{Law_2018_ECCV}
Law, H., Deng, J.: Cornernet: Detecting objects as paired keypoints. In: ECCV
  (September 2018)

\bibitem{li2019gs3d}
Li, B., Ouyang, W., Sheng, L., Zeng, X., Wang, X.: Gs3d: An efficient 3d object
  detection framework for autonomous driving. In: CVPR (2019)

\bibitem{Liang_CVPR_2019}
Liang, M., Yang, B., Chen, Y., Hu, R., Urtasun, R.: Multi-task multi-sensor
  fusion for 3d object detection. In: CVPR (2019)

\bibitem{Lin2016}
Lin, T., Doll{\'{a}}r, P., Girshick, R.B., He, K., Hariharan, B., Belongie,
  S.J.: Feature pyramid networks for object detection. CoRR
  \textbf{abs/1612.03144} (2016)

\bibitem{Lin+17}
Lin, T., Goyal, P., Girshick, R.B., He, K., Doll{\'{a}}r, P.: Focal loss for
  dense object detection. CoRR  \textbf{abs/1708.02002} (2017)

\bibitem{Liu_2018_DetSurvey}
Liu, L., Ouyang, W., Wang, X., Fieguth, P.W., Chen, J., Liu, X.,
  Pietik{\"{a}}inen, M.: Deep learning for generic object detection: {A}
  survey. CoRR  \textbf{abs/1809.02165} (2018)

\bibitem{Liu+19}
Liu, L., Lu, J., Xu, C., Tian, Q., Zhou, J.: Deep fitting degree scoring
  network for monocular 3d object detection. CoRR  \textbf{abs/1904.12681}
  (2019)

\bibitem{Liu2016}
Liu, W., Anguelov, D., Erhan, D., Szegedy, C., Reed, S., Fu, C.Y., Berg, A.C.:
  Ssd: Single shot multibox detector. In: ECCV (2016)

\bibitem{liu2020smoke}
Liu, Z., Wu, Z., Tóth, R.: Smoke: Single-stage monocular 3d object detection
  via keypoint estimation. CoRR  \textbf{abs/2002.10111} (2020)

\bibitem{Manhardt_2019_CVPR}
Manhardt, F., Kehl, W., Gaidon, A.: Roi-10d: Monocular lifting of 2d detection
  to 6d pose and metric shape. In: CVPR (2019)

\bibitem{Mousavian_2017_CVPR}
Mousavian, A., Anguelov, D., Flynn, J., Kosecka, J.: 3d bounding box estimation
  using deep learning and geometry. In: CVPR (July 2017)

\bibitem{Murthy_17_ICRA}
Murthy, K.J., Krishna, S.G., Chhaya, F., Krishna, M.K.: Reconstructing vehicles
  from a single image: Shape priors for road scene understanding. In: ICRA
  (2017)

\bibitem{Pillai_2019_ICRA}
Pillai, S., Ambrus, R., Gaidon, A.: Superdepth: Self-supervised, super-resolved
  monocular depth estimation. In: ICRA (2019)

\bibitem{qin2019monogrnet}
Qin, Z., Wang, J., Lu, Y.: Monogrnet: A geometric reasoning network for 3d
  object localization. In: (AAAI) (2019)

\bibitem{Redmon2016}
Redmon, J., Divvala, S., Girshick, R., Farhadi, A.: You only look once:
  Unified, real-time object detection. In: CVPR (June 2016)

\bibitem{Redmon_2017_CVPR}
Redmon, J., Farhadi, A.: Yolo9000: Better, faster, stronger. In: CVPR (2017)

\bibitem{Ren+15}
Ren, S., He, K., Girshick, R., Sun, J.: Faster {R-CNN}: Towards real-time
  object detection with region proposal networks. In: NIPS (2015)

\bibitem{Roddick18}
Roddick, T., Kendall, A., Cipolla, R.: Orthographic feature transform for
  monocular 3d object detection. CoRR  \textbf{abs/1811.08188} (2018)

\bibitem{RotPorKon18a}
Rota~Bul\`o, S., Porzi, L., Kontschieder, P.: In-place activated batchnorm for
  memory-optimized training of {DNN}s. In: CVPR (2018)

\bibitem{shi2018pointrcnn}
Shi, S., Wang, X., Li, H.: Pointrcnn: 3d object proposal generation and
  detection from point cloud. In: CVPR (2019)

\bibitem{Shin_arxiv_18}
Shin, K., Kwon, Y.P., Tomizuka, M.: Roarnet: {A} robust 3d object detection
  based on region approximation refinement. CoRR  \textbf{abs/1811.03818}
  (2018)

\bibitem{simonelli2019disentangling}
Simonelli, A., Rota~Bul{\`o}, S., Porzi, L., L{\'o}pez-Antequera, M.,
  Kontschieder, P.: Disentangling monocular 3d object detection. In: ICCV
  (2019)

\bibitem{wang2019pseudo}
Wang, Y., Chao, W.L., Garg, D., Hariharan, B., Campbell, M., Weinberger, K.:
  Pseudo-lidar from visual depth estimation: Bridging the gap in 3d object
  detection for autonomous driving. In: CVPR (2019)

\bibitem{Wang_arxiv_2019}
Wang, Z., Jia, K.: Frustum convnet: Sliding frustums to aggregate local
  point-wise features for amodal 3d object detection. CoRR
  \textbf{abs/1903.01864} (2019)

\bibitem{yang2019pseudopp}
You, Y., Wang, Y., Chao, W.L., Garg, D., Pleiss, G., Hariharan, B., Campbell,
  M., Weinberger, K.Q.: Pseudo-lidar++: Accurate depth for 3d object detection
  in autonomous driving. CoRR  \textbf{abs/1906.06310} (2019)

\bibitem{Zia_2014_CVPR}
Zia, M.Z., Stark, M., Schindler, K.: Are cars just 3d boxes? {J}ointly
  estimating the 3d shape of multiple objects. In: CVPR (2014)

\end{thebibliography}
